# LMF Reloaded

**Laurent Romary[1,3,4], Mohamed Khemakhem[1,2,3], Fahad Khan[8], Jack Bowers[1,6,7], Nicoletta Calzolari[8], Monte George[5], Mandy Pet[5], Piotr Bański[9]**

1. Inria-ALMAnaCH - Automatic Language Modelling and ANAlysis & Computational Humanities
2. UPD7 - Université Paris Diderot - Paris 7
3. CMB - Centre Marc Bloch
4. BBAW - Berlin-Brandenburg Academy of Sciences and Humanities
5. ANSI-  American National Standards Institute
6. EPHE - École Pratique des Hautes Études
7. ÖAW - Austrian Academy of Sciences
8. CNR-ILC - Istituto di Linguistica Computazionale "Antonio Zampolli"
9. IDS - Institut für Deutsche Sprache

## Abstract

The Lexical Markup Framework (LMF) or ISO 24613 [1] is a *de jure* standard which constitutes a framework for modelling and encoding lexical information both in retrodigitised print dictionaries as well as in NLP lexical databases. An in-depth review is currently underway within the standardisation sub-committee, ISO-TC37/SC4/WG4 with the goal of creating a more modular, flexible and durable follow up to the original LMF standard published by ISO in 2008. In this paper we will showcase some of the major improvements which have so far been implemented in the new version of LMF.

## 1. Introduction

The previous version of LMF, published by ISO in 2008 [1] offered a framework for modelling, publishing and sharing lexical resources with a special focus on requirements arising from the domain of Natural Language Processing (NLP).  Due to the potential richness and the multi-layered nature of linguistic descriptions in lexical resources the LMF meta-model ended up taking on a great deal of complexity in its attempt to reflect these various different linguistic facets. At the same time key areas of linguistics such as etymology (and diachronic lexical information in general) were not covered at all. Finally, the recommended serialisation for LMF was not clearly compatible with other leading markup standards, namely the TEI [3].

For these, and other reasons it was decided that the standardisation sub-committee, ISO-TC37/SC4/WG4 should review the LMF meta-model in order to create a new version of the standard which would address all of these issues. This new version of LMF will constitute a multi-part standard consisting of seven modules with the possibility of further extensions. Importantly the new version of LMF will be backwards compatible with the 2008 version. In the following section we will describe each package of the revised standard.

## 2. Abstract Modelling

In keeping with the fundamental conceptual modelling principles which have been decided on by ISO-TC37/SC4/WG4, the proposed model has been decoupled from any single serialisation format, although two potential serialisations of the meta-model constitute parts iv and v of the standard (TEI and LBX respectively). As a result, three major improvements have been carried out: restructuring; enrichment and simplification. Each are discussed below.

### a. Restructuring

Although the previous version of the standard reflected some separation among packages touching on different linguistic levels of description, the differentiation lacked a sufficient level of modularity. A user of the standard had to get the standard as a whole package where he could be interested in just specific parts of it. The current version of the standard is much more modular and has been split into the following seven parts:

i. ISO 24613-1 - Core model: defines basic classes required to model a baseline lexicon

ii. ISO 24613-2 - Machine Readable Dictionaries (MRD) model: contains components providing deeper specification of lexical description encapsulated within the core model. *Form* is for instance differentiated into *Related Form*, *Word Form, Stem* and *Word Part*

iii. ISO 24613-3 - Diachrony-Etymology: categories related to word and meaning origin and change are defined

iv. ISO 24613-4 - TEI serialisation: represents a first serialisation of the first three parts based on a restricted version of the Text Encoding Initiative (TEI) guidelines [2]

v. ISO 24613-5 - LBX serialisation: a second serialisation is formalised here using Language Base Exchange (LBX)

vi. ISO 24613-6 - Syntax and Semantics: semantic and syntactic components are gathered in this extension to be revised and integrated with the first three parts of the standard

vii. ISO 24613-7 - Morphology: morphology package will be defined in a separate part of the standard and will also be interconnected with the first three parts of the standard

The restructuring comes along with a revision of class membership; for example, the *Lemma* class, which was previously based in the MRD part is now part of the Core Module as it is a fundamentally essential part of a lexicon.

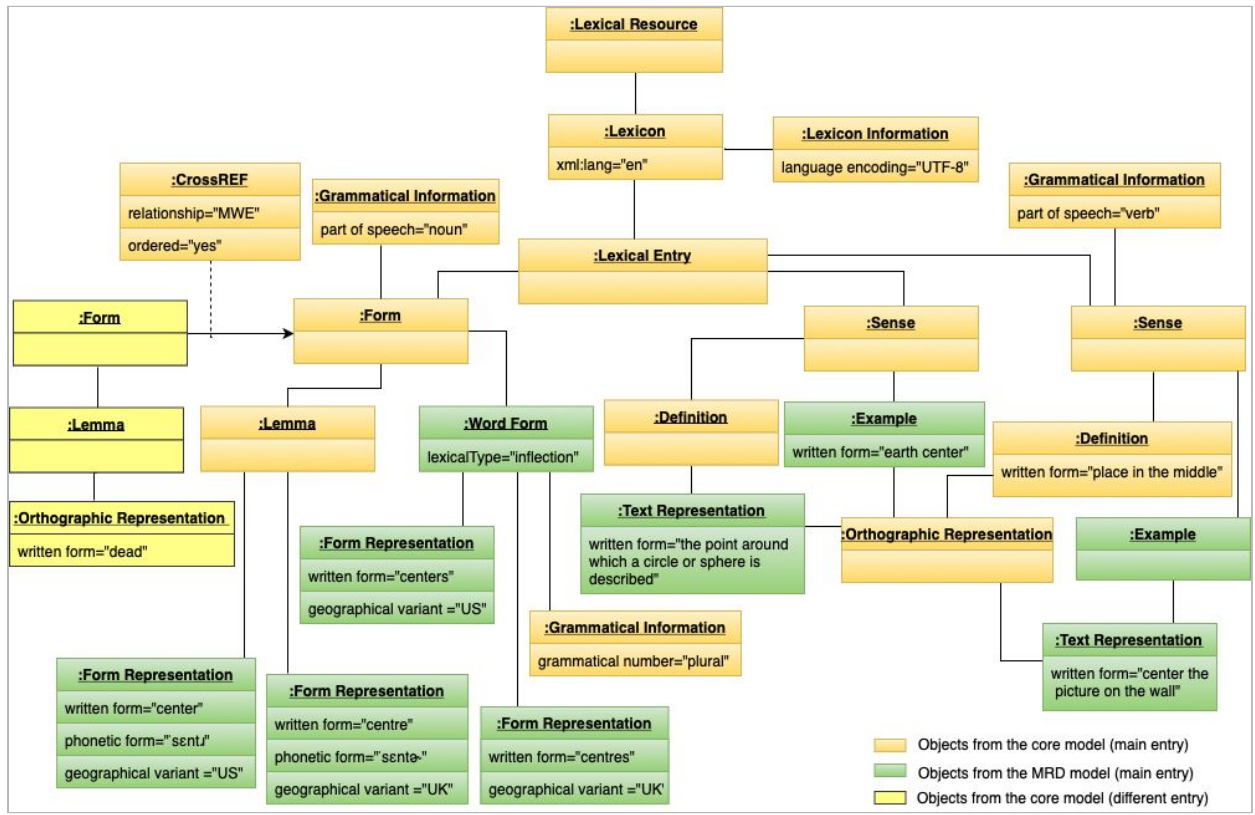

**Figure 1: Example of the lexical entry "Center" encoded using the core (ISO 24613-1) and MRD (ISO 24613-2) metamodels**

### b. Enrichment

New information has been introduced to describe essential aspects of lexical information such as *Bibliography*. Such information is required to specify references for some usages, definitions, examples, etc. Therefore the new class is kept multi-functional to be used in case of need as determined by the editor of a lexicon. Additionally, the differentiation of Orthographic Representation into Form Representation and Text Representation has been designed to enable more precision in the encoding of written forms touching respectively Sense and Form sub-classes.

### c. Simplification

The emphasis on abstraction and modularisation has also led to a series of major simplifications affecting nearly every class of the meta-model. One key feature which is been newly introduced is the CrossREF class which is a pointing/mapping mechanism that can be used to model a wide array of lexical features and relationships such as semantic relations, cross references, related entries and others within the meta-model. As a result some classes (e.g. *List of Components* and *Component)* whose features have been taken on, in part, by *CrossREF* have been removed altogether. Figure 1 illustrates the simplicity of the new mechanism used to model Multi Word Expressions (MWE) previously represented by classes which are now obsolete.

## 3. More Coverage

One of the main intentions behind the new version of the standard is to provide increased coverage of the type of information that can be encoded by the model. To this end a completely new meta-model covering etymological and diachronic information is proposed in (ISO 24613-3). This new module extends the core LMF and MRD metamodels adding the following key classes (among others):

- **Etymon** and **Cognate**: subclasses of the core class Lexical Entry which are used in describing the diachrony of other lexical entries. Etymons are lexical entries from which another lexical entry is derived (a historical form or sense), and Cognates are lexical entries in related languages which share a common ancestor with a given aspect of a lexical entry.
- **EtyLink**: subclass of CrossREF which is used to link one or more temporal stages of one or more aspects of a lexical entry (i.e. sense, phonetic properties, etc.)
- **Etymology**: describes the history of a lexical entry or other element by being associated with an ordered series of EtyLink instances. An Etymology instance can be recursive and typed to define the changes undergone according to any number of linguistic processes (e.g. borrowing, inheritance, metaphor, metonymy,  etc.)
- **Date**: defines specific or relative temporal information associated with some aspect of an etymology or its components

We illustrate the new etymology module with an example, an etymology for the word *center* which we have taken from Klein [4]:

center, centre, n. — F. *centre*, fr. L. *centrum*, fr. Gk. χέντρον, 'point, prickle, spike, ox goad, point round which a circle is described', from the stem of χέντειν, 'to prick, goad', whence also χέντωρ, 'a goader, driver', χεστός (for *χεντ-τός), 'embroidered', χέστρᾱ, 'pickax', χοντός, 'pole', fr. I.-E. base *ḱent-, 'to prick', whence also Bret. *kentr*, OIr. *cinteir*, 'a spur', OHG. *hantag*, 'sharp, pointed', Lett. *sîts*, 'hunter's spear', *situ*, *sist*, 'to strike', W. *cethr*, 'nail'. Cp. **centrifugal, centripetal, concentrate, eccentric, Dicentra, paracentesis.** Cp. also **cestrum, cestus,** 'girdle', **kent,** 'a pole', **quant,** 'a pole'.
Derivatives: *center*, *centre*, intr. and tr. v., *center-ing*, *centr-ing*, *centre-ing*, n.

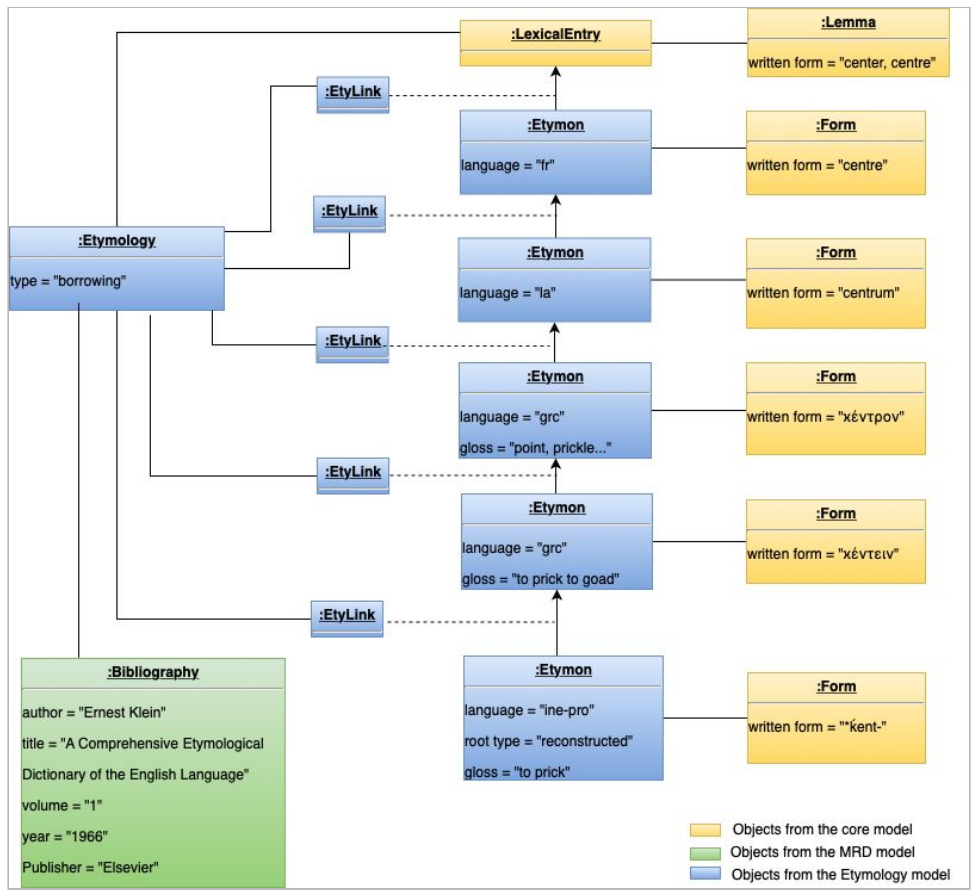

**Figure 2: Example of modelling Etymological information following ISO 24613-3 meta-model, together with the entry being modelled from [4]**

In Figure 2 we present an LMF model of a part of the Klein entry, namely, the portion that describes the borrowing of the word *center* from French, and its history prior to that.

Note that for reasons of space we haven't shown the ordering of the Etylinks in the diagram nor specified the types of link (specifying for instance if an etymological link between two elements represents an etymological borrowing or inheritance). This is however an important aspect of etymological encoding and should be included in LMF resources describing the etymologies of lexical entries.

## 4. Connection with Leading Standards

The Text Encoding Initiative (TEI) is a standard which is widely adopted among lexicographers. The new version of LMF contains a TEI serialisation which aims to make both standards fully compatible following the vision presented by Romary in [3]. This serialization has the benefit of being able to leverage the knowledge and make use of the established practices of the TEI community in dealing with the representation of a wide array of lexicographic issues with which LMF is also concerned.

```xml
<entry>
    <form type="lemma" xml:id="center_form">
        <orth>center</orth>
        <pron>ˈsɛntɹ</pron>
        <gramGrp>
            <pos>noun</pos>
        </gramGrp>
        <usg type="geo">U.S</usg>
        <form type="variant">
            <orth>centre</orth>
            <usg type="geo">U.K</usg>
            <pron>ˈsɛntɚ</pron>
        </form>
    </form>
    <form type="inflected">
        <orth>centers</orth>
        <usg type="geo">U.S</usg>
        <gramGrp>
            <number>plural</number>
        </gramGrp>
    </form>
    <form type="inflected">
        <orth>centres</orth>
        <usg type="geo">U.K</usg>
        <gram type="number">plural</gram>
    </form>
    <sense>
        <def>the point around which a circle or sphere is described</def>
        <cit type="example">
            <quote>earth center"</quote>
        </cit>
    </sense>
    <sense>
        <gramGrp>
            <pos>verb</pos>
        </gramGrp>
        <def>place in the middle</def>
        <cit type="example">
            <quote>center the picture on the wall</quote>
        </cit>
    </sense>
    <re type="multiWordExpression">
        <form>
            <seg corresp="#dead_form" n="1">dead</seg>
            <seg corresp="#center_form" n="2">center</seg>
        </form>
    </re>
</entry>
```

**Figure 3: Encoding example following LMF's TEI serialisation (ISO 24613-4)**

The TEI guidelines offer a great degree of freedom for encoding lexical information. However in some cases, such freedom comes at the cost of an excess of variability in how users choose to represent certain features. Therefore in this serialization, we have sought to constrain that flexibility in line with similar initiatives, namely TEI Lex-0 [5,6,7]. In Figure 3 we show how the components in Figure 1 can be serialised using TEI elements. The development of a list[1], gathering serialisation examples provided by the community and checked by the ISO experts, along with a schema specification[2] is underway.

## 5. Conclusion

In this work we have presented the measures we followed to remedy the deficiencies noted in LMF after years of release. The changes, being in some cases important, will bring more flexibility and interoperability to the standard. The in-depth structure review along with the enriching new modules will be great assets for the current users of the standards and hesitant lexicographers to adopt the standard as they could benefit not only from the advantages of a de jure standard like ISO 24613 but also to have efficient modelling and serialisation alternatives.

[1] https://github.com/DARIAH-ERIC/lexicalresources/blob/master/Schemas/LMFinTEI%20Specification/examplesLMFinTEI.xml

[2] https://github.com/DARIAH-ERIC/lexicalresources/blob/master/Schemas/LMFinTEI%20Specification/LMFinTEIspec.html